\documentclass[10pt,twocolumn]{article}
\usepackage[a4paper,margin=20mm,top=22mm,bottom=22mm]{geometry}
\usepackage[utf8]{inputenc}
\usepackage[T1]{fontenc}
\usepackage{times}
\usepackage{microtype}
\usepackage{booktabs}
\usepackage{array}
\usepackage{enumitem}
\usepackage{xcolor}
\usepackage{url}
\usepackage[colorlinks=true,linkcolor=black,citecolor=black,urlcolor=blue!50!black]{hyperref}
\usepackage{titlesec}
\usepackage{caption}
\usepackage{graphicx}
\usepackage{fancyhdr}

\titleformat{\section}{\large\bfseries}{\thesection}{0.6em}{}
\titleformat{\subsection}{\normalsize\bfseries}{\thesubsection}{0.6em}{}
\titlespacing{\section}{0pt}{10pt}{4pt}
\titlespacing{\subsection}{0pt}{7pt}{3pt}
\setlist{nosep,leftmargin=*}
\newcommand{\code}[1]{\texttt{\small #1}}

\title{\textbf{Measure Before You Train:\\ Typed Decision Models vs.\ a Classical Baseline vs.\ a Generative LLM\\ for Routing a Personal Multilingual Inbox}}
\author{Federico Lucca\\
\small Independent researcher, Trento, Italy\\
\small \texttt{federico.lucca@gmail.com}}
\date{\small Preprint v2.2, November 2026 — arXiv:2610.32955}

\begin{document}
\twocolumn[
\maketitle
\begin{abstract}
\noindent
Personal email agents decide, for every incoming message, its type and urgency and whether to draft a reply, create a task or ignore it. In production these decisions are delegated to a generative LLM or to rules, and their error rate is rarely measured: the stored labels were written by the system itself. We describe Foyer, a self-hosted personal inbox agent, and a measurement protocol---stratified sampling, a frozen snapshot of the model input, thread-grouped splits, secret redaction, and human labels collected in the agent's own review loop---that produces ground truth before any training. On 231 real emails in Italian and English we compare eleven systems: the stored production type (keyword rules and an earlier LLM classifier), a TF-IDF + logistic-regression baseline, local typed decision models (Tev1 0.8B and 4B, Nimble 9B, Jeva-4B), two hosted ones (Bespoke Nimble v4-max and TypeSafe Jev) and three hosted generative LLMs (Qwen2.5-7B, Gemma~4 26B-A4B and 31B). Four results. First, the stored production type scores 28\% against a 57\% majority class. Second, with about two hundred labels a logistic regression has the highest point accuracy on every field (type $0.77\pm0.02$ over five seeds of the split, urgency 0.74, needs-reply 0.83, is-noise 0.90) at a latency three orders of magnitude lower, though it never predicts the two rarest types; but on every field at least one hosted model is statistically indistinguishable from it (exact McNemar, Holm-corrected)---Gemma~4 on type (0.74), Nimble v4-max on urgency and is-noise (0.71, 0.86). Third, a generative model recovers part of the sender's relationship to the owner from the text (0.60); inheritance from the contact is right 98.6\% of the time on the 55--63\% of emails whose sender was seen before (0.86 on all emails, with a majority fallback): the relationship belongs in the contact graph, not in the prompt. Fourth, per-email accuracy is not the product metric: composed into actions, the classifier ignores 18\% of the emails that needed a reply or a task, against 4\% for the best generative model; a cost-chosen threshold cuts that rate to 10\% (7/68), indistinguishable from the hosted decision model's. On the unbiased corpus no cost difference among the leading systems is significant; on the adversarial corpus the thresholded classifier is the cheapest system for any missed-email price above 4. A third, adversarially sampled round (722 emails in total) with eleven pre-registered hypotheses confirms seven and refutes four. Two blind annotators agree with the reference at $\kappa=0.81$ and $0.86$ on message type. We release the protocol, the harness and the pre-registration; the emails stay private.
\end{abstract}
\vspace{1.5em}
]

\section{Introduction}

A growing share of what personal AI agents do is not generation but decision: which project does this email belong to, is it urgent, does it need a reply, is it noise. Generative LLMs answer these questions by producing text that is then parsed; \emph{typed decision models}, introduced with TypeSafe's Jev in September 2026~\cite{jev} and followed within weeks by open models (Nimble~\cite{nimble}, Tev1, Clef, Strands Decider) and a local endpoint in Ollama~\cite{ollama}, answer them directly with a probability distribution over caller-defined options, in one forward pass and without free text.

Whether this new class of models is good enough to replace the generative step in a personal agent is an empirical question, and it cannot be answered with the data such agents typically keep: the stored labels are the model's own output. Early evidence is mixed. Classical classifiers trained on a few thousand records outperform Jev on network-traffic classification~\cite{traffic}; a community benchmark reports a plain 9B instruct model ahead of every decision model on triage tasks; the vendor's own benchmark places the open models within three points of Jev~\cite{ollama}. None of it concerns personal, multilingual correspondence.

This paper makes three contributions.
\begin{enumerate}
\item A \textbf{measurement protocol} for a production email agent that yields human ground truth without changing the system: stratified sampling by predicted class, a frozen snapshot of the exact model input, deterministic splits keyed on record ids, and labelling in the agent's own interface.
\item A \textbf{comparison of eleven systems} on that ground truth---the stored production type, a classical baseline, local and hosted typed decision models, hosted generative LLMs---on message type, urgency, three agent switches (draft a reply, create a task, ignore) and project assignment, with calibration, latency and cost.
\item A \textbf{decomposition} of the agent's labels into what is learnable from the text of one message (type, urgency, the three switches) and what one message carries only in part (the sender's relationship to the owner: 0.60 for the best generative model), with a measured alternative for the latter: inheritance from the contact graph.
\end{enumerate}

\section{Related Work}

\paragraph{Typed decision models.}
Jev~\cite{jev} defined the interface: a state, a set of typed questions (choice, yes/no, score), and a distribution per question. Open models followed: Bespoke Nimble~9B, fine-tuned from Qwen3.5-9B with LoRA on answer tokens only and released with its contrastively curated data and recipe~\cite{nimble}---its held-out comparison places the untuned 9B base at 66\% against 90\% for Nimble and 93\% for Jev, and its evaluation notes insist that models be compared only on the same examples under the same procedure, which we adopt; Together's Tev1 4B/0.8B, Cloudflare's Clef, Amazon's Strands Decider and Perplexity's pplx-decider; Ollama~0.35 exposed them locally through \code{/v1/systemone}~\cite{ollama}. The LLM-as-Jev framework shows that any causal LLM can be scored under the same convention~\cite{llmasjev}. Jeva~\cite{jeva} encodes the request and each option separately, with an explicit null outcome, permutation invariance, scaling to thousands of candidates and reuse of option encodings across requests---a design aimed at two failures we also see in project assignment: over-use of \emph{none} and re-encoding a fixed candidate set on every call; on this corpus the candidate-set reuse holds, while null awareness is an open-set property that our fixed-set comparison does not test (Section~\ref{sec:results}). An early evidence audit found 29 arXiv papers mentioning typed decision models within nine days of the Jev release and proposes an evaluation checklist~\cite{audit}, which we follow; a broader benchmark places Jev against nineteen general-purpose LLMs~\cite{bench}, and a data-driven analysis maps its functionality, applications and ecosystem~\cite{wild}.

\paragraph{Decision models vs.\ classical classifiers.}
On network-flow classification, Random Forest and Extra Trees trained on 8{,}000 records beat Jev in every test week~\cite{traffic}; a community test of 14 decision models on 56 triage tasks found a plain Qwen3.5-9B ahead of all of them. We take the classical baseline as the cost floor every neural option must beat.

\paragraph{Email classification and personal agents.}
Email classification predates LLMs by three decades: rule learning for mail filing~\cite{cohen96}, folder prediction on the Enron corpus~\cite{klimt04,bekkerman04}, reply prediction---``does this message need a reply''---and attachment prediction as the first ``intelligent email'' tasks~\cite{dredze08}, the former being exactly our needs-reply switch, and the observation that users need a handful of broad categories rather than many folders~\cite{grbovic14}, which is what our six types and three switches are. That work trained per-user classifiers on a few hundred labelled messages, which is also where our baseline sits. What is new is the alternative: zero-shot decisions from a language model, generative or typed, without any per-user data---and the question of whether, for one inbox, that alternative beats a classifier trained on labels the user can produce in an afternoon. Privacy motivates the local setting: keeping the model where the data are is the premise of federated and on-device learning~\cite{mcmahan17}, and personal correspondence is the canonical example of data that should not leave the device.

\section{The System}\label{sec:system}

Foyer is a self-hosted personal inbox agent in daily use: it ingests the owner's mail accounts, decides for each message what it is and what should happen to it, assigns it to a project, and runs small background agents on the result. For this study only the routing decisions matter, and at the time of the measurements they were produced as follows. Obvious noise (newsletters, bot senders, the owner's own mail) was filtered by lists and rules. \emph{Type} and \emph{urgency} were assigned at ingest by a keyword heuristic (Appendix~B.5): type is \emph{offer}, \emph{request} or \emph{update} if one of a handful of Italian/English keywords appears in subject or snippet, else \emph{other}; urgency is \emph{high} or \emph{medium} on a list of urgency words, else \emph{low}. The heuristic can never produce \emph{client}, \emph{spam} or \emph{newsletter}. The stored \emph{type} field, however, also holds values written earlier by an LLM batch classifier whose classes included \emph{client}, \emph{spam} and \emph{newsletter}; on the 231 v2 emails it reads \emph{client} 32 times. What we evaluate as the production baseline is that stored field, the \emph{stored production type}, i.e.\ what the agent actually displayed, not the heuristic alone. \emph{Project assignment} was the only decision taken by a generative LLM, prompted with a group of messages and the list of existing projects. Every decision is logged with a flow id, which gives provenance for the error analysis.

At the time of the study the LLM backend was a hosted gateway with a local Ollama model as fallback; the local model had been demoted after timing out on CPU. The study therefore measures the production system as it actually ran, not as its documentation describes it: reading the code rather than the README revealed both that the local model was no longer primary and that type and urgency were never produced by a model at all. The pipeline above is the one measured here; after the measurements it was changed (Section~\ref{sec:changes}).

\section{Measurement Protocol}

The stored classification field is written by the model; counting it says nothing about error. We added a \code{dataset\_email} command that:
\begin{itemize}
\item \textbf{samples stratified by predicted type} (seven classes: \emph{request, update, offer, client, other, spam, newsletter}), because 39\% of stored labels are \emph{other} and 94\% of urgencies \emph{low}; uniform sampling would teach the prior, not the content;
\item \textbf{freezes a snapshot} of exactly what the model sees (sender, subject, body truncated to 1{,}500 characters, date, channel) next to the model's prediction, so later re-runs cannot change the example;
\item \textbf{assigns splits deterministically by thread}, stratified by type: a first version keyed on \code{sha1(id)} put 11 threads across splits and gave 4 test emails a near-duplicate in train, so the unit is the thread (normalised subject + sender key), and, while the test split is below 60 emails, the primary protocol is 5-fold thread-grouped cross-validation with the baseline trained out-of-fold;
\item \textbf{redacts secrets} in the frozen body (passwords, tokens, card and account numbers) before storage---a pilot export contained a VPN password in clear text;
\item \textbf{collects human labels in the interface}: type, area, project, urgency, needs-reply, action-within-48h, is-noise;
\item reports, before any training, the only number that matters: how often the production model was right;
\item \textbf{pre-registers} the hypotheses of each new labelling round: for the third round, eleven hypotheses with thresholds and a verification script were committed before the labelled file was opened (one post-hoc change: a format fix in the script, not a threshold).
\end{itemize}
In the first round the id-keyed split placed 11 of 92 threads across splits and gave 4 test emails a near-duplicate in train; the second round was split by thread and stratified by type (train 159 / validation 39 / test 33), with one inherited thread still crossing. Because the fixed test split holds no \emph{high} and one \emph{offer}, every number in this paper comes from thread-grouped cross-validation (trained baseline) or from all 231 rows (zero-shot systems); the fixed split is kept in the repository for future rounds.

\section{Experimental Setup}\label{sec:setup}

\paragraph{Data.} Three labelling rounds on the owner's production database. Rounds 1--2 were sampled stratified by the stored production type: 119 emails on 20 October 2026 (4 skipped as ambiguous) and 120 on 21 October; eight rows marked \emph{ignore} (private family mail, empty bodies) are excluded, leaving the \textbf{v2 corpus: 231 emails in 207 threads}, our unbiased reference. Round 3 (500 emails, 491 kept) was sampled \emph{adversarially}: all 118 emails of the unlabelled pool in which Nimble or Gemma~4 saw \emph{offer} or \emph{spam} (G1), 125 on which the models disagreed on action-48h (G2), and 257 at random stratified on the baseline's prediction (G3). The \textbf{v3 corpus (722 emails, 698 threads)} is therefore a stress test built on the LLMs' presumed errors, not a sample of the natural inbox; system comparisons are reported on v2, and v3 is used for the pre-registered hypotheses, the contact-graph result and the action analysis. In the first round two model annotators proposed type and urgency and the owner confirmed or corrected each row in a spreadsheet that showed sender, subject, 700 characters of body and both proposals but not the production prediction (110/115 types and 104/115 urgencies survived review); in the second round the owner labelled from the model proposal with the same interface, then the full set was revised jointly with an assistant to apply the final label definitions below. The ground truth is therefore a single human reviewer anchored on model proposals; two blind annotators relabelled 60 v2 emails from scratch to measure how far it is shared (Section~\ref{sec:limits}). Class distribution in Table~\ref{tab:data}.

\paragraph{Labels.} Label names are given in English throughout; the questions and class definitions shown to the models were in Italian (verbatim in Appendix~B). Two axes were separated after the first round, because the stored \emph{type} mixed what a message says with who sent it. \emph{Type} (six classes: request, update, offer, newsletter, spam, other) describes the message. \emph{Relationship} (client, supplier, partner, family, institution, service, monitoring, self) describes the sender and is assigned per contact; \emph{client} is someone who gives the owner work, an actual paying client rather than a prospective one (the definition the labels follow; Appendix~C). The first-round guideline used a broader wording (any interaction that can lead to a deal); the five first-round emails labelled \emph{client} under it are superseded by the v2 labels, and no reported result depends on that wording. \emph{Urgency} has three levels. Three booleans are the agent's switches: \emph{needs-reply} (the agent drafts an answer), \emph{action-48h} (the agent creates a task) and \emph{is-noise} (the agents ignore the message; it is kept only as context). \emph{Project} is the project the email belongs to as of October 2026; 26 rows are deliberately empty (``none'') because the project-manager agent has not proposed one yet.

\begin{table}[t]
\centering\small
\caption{Labelled corpus (231 emails after exclusions).}
\label{tab:data}
\resizebox{\columnwidth}{!}{%
\begin{tabular}{lr@{\hspace{1.2em}}lr}
\toprule
Type & n & Relationship & n \\
\midrule
update & 132 & service & 115 \\
newsletter & 53 & client & 69 \\
request & 20 & self & 18 \\
other & 18 & monitoring & 14 \\
offer & 5 & supplier & 8 \\
spam & 3 & family / partner / institution & 7 / 6 / 2 \\
\midrule
Urgency low / medium / high & 161 / 59 / 11 & needs-reply yes & 60 \\
Projects (incl.\ none) & 16 + 1 & action-48h yes / is-noise yes & 20 / 119 \\
\bottomrule
\end{tabular}}
\end{table}

\paragraph{Systems.} \emph{Stored production type}: the type and urgency stored in the production database, written by the keyword heuristic of Appendix~B.5 and, for part of the emails, by the earlier LLM batch classifier (Section~\ref{sec:system}); it has no output for the booleans and holds \emph{client} for 32 rows, wrong by construction after the label change. \emph{Classical baseline}: TF-IDF word 1--2-grams (20{,}000 features), logistic regression with balanced class weights, 5-fold cross-validation grouped by thread (207 threads), the only trained system. Its figures in the tables come from one split (seed 0); over five seeds of the split it scores $0.771\pm0.021$ on type ($0.733\pm0.010$ urgency, $0.866\pm0.020$ needs-reply, $0.923\pm0.012$ action-48h, $0.917\pm0.010$ is-noise). Adding character 3--5-grams gives $0.785\pm0.014$ on type; we report it as a note, not as the baseline. \emph{Typed decision models}: Tev1 0.8B on CPU (i5-1245UE), Tev1 4B and Nimble 9B (Q8) on a Kaggle T4 via Ollama \code{/v1/systemone}, Jeva-4B~\cite{jeva} in-process on the same GPU, and Bespoke Nimble v4-max and TypeSafe Jev (jev-1.14.0) through their APIs. \emph{Generative LLMs}: Qwen2.5-7B-Instruct, Gemma~4 26B-A4B and Gemma~4 31B through OpenRouter, JSON mode, temperature 0. All systems received the same Italian question and one-line class definitions (Appendix~B), and the typed decision models the same question format: \code{choice} for type and urgency, \code{choice} yes/no with one definition per option for the three switches (Jev, Nimble v4-max and Jeva-4B also as \code{noul}, Table~\ref{tab:bool}); the body was stripped of markup and URLs and truncated to 1{,}500 characters. Zero-shot systems were run once on all 231 rows. The hosted Gemma~4 26B is not deterministic at temperature 0 (0.774, 0.800, 0.791 on the first-round corpus; on type here, 0.740 and 0.749 on two runs); hosted figures are single-run estimates.

\paragraph{Metrics.} Accuracy and macro-F1 against the majority class; for systems that return probabilities, mean confidence on correct vs.\ incorrect decisions; latency as median per decision; cost per 1{,}000 emails from provider usage. With 231 rows one email is 0.43 accuracy points; \emph{offer} (5) and \emph{spam} (3) make macro-F1 on type unstable, so for type we read accuracy and the F1 of the four large classes.

\section{Results}\label{sec:results}

\begin{table}[t]
\centering\small
\caption{Message type (6 classes) and urgency (3 levels), 231 emails. B = TF-IDF + logistic regression, thread-grouped 5-fold CV; all other systems zero-shot. 95\% bootstrap intervals on accuracy (2,000 resamples of threads). B is one split (seed 0); over five seeds, type $0.771\pm0.021$. Tests against B: Table~\ref{tab:tests}. Latency is the median per decision.}
\label{tab:type}
\resizebox{\columnwidth}{!}{%
\begin{tabular}{lrrrrr}
\toprule
 & \multicolumn{2}{c}{Type} & \multicolumn{2}{c}{Urgency} & \\
System & Acc.\ [95\% CI] & M-F1 & Acc.\ [95\% CI] & M-F1 & Latency \\
\midrule
Majority class & 0.571 & 0.121 & 0.697 & 0.274 & -- \\
Stored production type & 0.277 [.22--.34] & 0.205 & 0.684 [.61--.75] & 0.291 & $<$1\,ms \\
B: TF-IDF + LogReg & \textbf{0.788} [.73--.85] & 0.474 & \textbf{0.736} [.67--.80] & \textbf{0.548} & $<$1\,ms \\
Tev1 0.8B, local CPU & 0.623 [.56--.69] & 0.284 & 0.251 [.19--.32] & 0.235 & 3.8\,s \\
Tev1 4B, local T4 & 0.688 [.62--.75] & 0.386 & 0.450 [.38--.52] & 0.379 & 0.35\,s \\
Nimble 9B Q8, local T4 & 0.619 [.55--.69] & 0.326 & 0.411 [.35--.48] & 0.357 & 0.58\,s \\
Jeva-4B, local T4 & 0.498 [.43--.57] & 0.275 & 0.264 [.20--.33] & 0.234 & 1.03\,s \\
Nimble v4-max, hosted & 0.701 [.64--.76] & 0.456 & 0.714 [.65--.78] & 0.541 & 0.50\,s \\
Jev 1.14, hosted & 0.727 [.66--.79] & 0.513 & 0.550 [.48--.62] & 0.422 & 0.22\,s \\
Qwen2.5-7B, hosted gen. & 0.571 [.50--.64] & 0.286 & 0.437 [.37--.51] & 0.393 & 0.48\,s \\
Gemma 4 26B-A4B, hosted gen. & 0.740 [.68--.80] & \textbf{0.542} & 0.628 [.56--.69] & 0.504 & 0.76\,s \\
Gemma 4 31B, hosted gen. & 0.732 [.67--.80] & 0.510 & 0.645 [.58--.72] & 0.497 & 0.72\,s \\
\bottomrule
\end{tabular}}
\end{table}

\begin{table}[t]
\centering\small
\caption{The three agent switches, 231 emails (accuracy / macro-F1). $^\dagger$Jev asked the switches as \code{choice} yes/no with the definitions given to Nimble; asked as \code{noul} (probability of yes) without option definitions, it scores 0.827 / 0.767, 0.892 / 0.592 and 0.675 / 0.655; Nimble v4-max asked as \code{noul} scores 0.788 / 0.670, 0.836 / 0.558 and 0.866 / 0.866. $^\ddagger$Same \code{choice} format; Jeva-4B asked as \code{noul} scores 0.775 / 0.694, 0.883 / 0.533 and 0.524 / 0.411.}
\label{tab:bool}
\resizebox{\columnwidth}{!}{%
\begin{tabular}{lccc}
\toprule
System & needs-reply & action-48h & is-noise \\
 & (yes 60) & (yes 20) & (yes 119) \\
\midrule
Majority class & 0.740 / 0.425 & 0.913 / 0.477 & 0.515 / 0.340 \\
B: TF-IDF + LogReg & \textbf{0.831} / 0.742 & \textbf{0.931} / \textbf{0.731} & \textbf{0.900} / \textbf{0.899} \\
Tev1 0.8B, local & 0.706 / 0.465 & 0.840 / 0.578 & 0.628 / 0.626 \\
Jeva-4B, local T4$^\ddagger$ & 0.792 / \textbf{0.761} & 0.805 / 0.587 & 0.732 / 0.725 \\
Nimble v4-max, hosted & 0.805 / 0.743 & 0.805 / 0.536 & 0.857 / 0.857 \\
Jev 1.14, hosted$^\dagger$ & 0.792 / 0.738 & 0.857 / 0.558 & 0.783 / 0.780 \\
Gemma 4 26B-A4B & 0.796 / 0.731 & 0.849 / 0.569 & 0.809 / 0.806 \\
Gemma 4 31B & 0.805 / 0.748 & 0.840 / 0.562 & 0.771 / 0.766 \\
\bottomrule
\end{tabular}}
\end{table}

\begin{table}[t]
\centering\small
\caption{Ablations of the trained baseline (thread-grouped 5-fold CV, same folds within the table; accuracy / macro-F1). ``blind'' = subject and body only, no sender.}
\label{tab:abl}
\resizebox{\columnwidth}{!}{%
\begin{tabular}{lccccc}
\toprule
Input / features & type & urgency & needs-reply & action-48h & is-noise \\
\midrule
TF-IDF, with sender & 0.74 / 0.38 & 0.71 / 0.41 & 0.84 / 0.78 & 0.90 / 0.47 & 0.89 / 0.89 \\
TF-IDF, blind & 0.70 / 0.28 & 0.71 / 0.40 & 0.83 / 0.76 & 0.89 / 0.47 & 0.87 / 0.87 \\
MiniLM-L6 (en), with sender & 0.55 / 0.26 & 0.66 / 0.46 & 0.73 / 0.67 & 0.84 / 0.52 & 0.81 / 0.81 \\
MiniLM-L12 multilingual, with sender & 0.69 / 0.43 & 0.62 / 0.40 & 0.76 / 0.71 & 0.83 / 0.50 & 0.84 / 0.84 \\
Majority class & 0.57 & 0.70 & 0.74 & 0.91 & 0.52 \\
\bottomrule
\end{tabular}}
\end{table}

\begin{table}[t]
\centering\small
\caption{Relationship inherited from other emails of the same contact (leave-one-out), vs.\ the majority class.}
\label{tab:rel}
\resizebox{\columnwidth}{!}{%
\begin{tabular}{lrrrr}
\toprule
Key & Covered & Acc.\ (covered) & M-F1 (covered) & Acc.\ (all, fallback) \\
\midrule
same address & 146/231 (63\%) & \textbf{0.986} & 0.953 & \textbf{0.857} \\
same domain & 173/231 (75\%) & 0.890 & 0.574 & 0.849 \\
majority (\emph{service}) & -- & -- & -- & 0.498 \\
\bottomrule
\end{tabular}}
\end{table}

\begin{table}[t]
\centering\small
\caption{Labels composed into actions (v2, policy ``noise wins''). False noise = emails needing a reply or task that were ignored, of 68.}
\label{tab:actions}
\resizebox{\columnwidth}{!}{%
\begin{tabular}{lcccc}
\toprule
System & Draft P / R & Task P / R & False noise & Useless tasks \\
\midrule
B: TF-IDF + LogReg & 0.80 / 0.47 & 0.71 / 0.25 & 12/68 (18\%) & 2/7 \\
Nimble v4-max & 0.65 / 0.61 & 0.15 / 0.25 & 8/68 (12\%) & 28/33 \\
Gemma 4 26B-A4B & 0.62 / 0.59 & 0.21 / 0.25 & \textbf{3/68 (4\%)} & 19/24 \\
Gemma 4 31B & 0.63 / 0.64 & 0.19 / 0.25 & 5/68 (7\%) & 21/26 \\
Tev1 0.8B & 0.31 / 0.07 & 0.35 / 0.30 & 25/68 (37\%) & 11/17 \\
\midrule
\multicolumn{5}{l}{\emph{+ relationship from contact and monitoring rule:} task recall B 0.40, Nimble 0.50, Gemma 26B 0.65} \\
\bottomrule
\end{tabular}}
\end{table}

\begin{table}[t]
\centering\small
\caption{Project assignment on v2 (231 emails), 16 projects + none; zero-shot models see only the project names.}
\label{tab:proj}
\resizebox{\columnwidth}{!}{%
\begin{tabular}{lrrr}
\toprule
System & Acc.\ (all) & Acc.\ (with project, n=205) & none P / R \\
\midrule
Majority (\emph{personal}) & 0.316 & -- & -- \\
B: TF-IDF + LogReg & \textbf{0.697} & \textbf{0.712} & 0.75 / 0.58 \\
Tev1 0.8B & 0.260 & 0.234 & 0.22 / 0.46 \\
Nimble v4-max & 0.303 & 0.259 & 0.13 / 0.65 \\
Gemma 4 26B-A4B & 0.329 & 0.278 & 0.13 / 0.73 \\
Gemma 4 31B & 0.468 & 0.468 & 0.21 / 0.46 \\
\bottomrule
\end{tabular}}
\end{table}

\begin{table}[t]
\centering\small
\caption{Project assignment on the adversarial v3 corpus (352 emails with a project), 18 projects + none, with a one-line description per project (truncated to 120 characters) vs.\ names only. Local models on a T4, hosted models through their APIs. Majority class 0.330. Jeva-4B with \emph{none} as an ordinary option (\code{allow\_null=False}); Jeva's native null (\code{allow\_null=True}) is an open-set setting and is reported separately: it answers \emph{none} on 229 of 352 emails (precision 0.05, recall 0.92); without an independent validation split to calibrate the null threshold, that result is uncalibrated and not comparable with the fixed 19-class rows.}
\label{tab:projdesc}
\resizebox{\columnwidth}{!}{%
\begin{tabular}{lcccc}
\toprule
 & \multicolumn{2}{c}{Descriptions} & \multicolumn{2}{c}{Names only} \\
System & Acc. & M-F1 & Acc. & M-F1 \\
\midrule
Nimble 9B Q8, local & 0.540 & 0.584 & 0.264 & 0.450 \\
Tev1 4B, local & 0.253 & 0.512 & 0.213 & 0.401 \\
Jeva-4B, local & 0.114 & 0.191 & 0.210 & 0.216 \\
Nimble v4-max, hosted & \textbf{0.682} & 0.626 & 0.281 & 0.475 \\
Gemma 4 26B-A4B, hosted & 0.545 & 0.596 & 0.287 & 0.526 \\
Jev 1.14, hosted & 0.619 & \textbf{0.635} & 0.270 & 0.521 \\
\bottomrule
\end{tabular}}
\end{table}

\textbf{The stored production type is worse than no classifier.} The \emph{type} field of every email in the system, populated by the keyword heuristic and in part by the earlier LLM classifier, reaches 0.28 accuracy against a 0.57 majority class; on urgency it sits just below the majority (0.68 vs 0.70). Because the sample was stratified on this field itself, this is not an unbiased estimate of its production accuracy, but a field that performs below ``always say the most common class'' on a sample built to be favourable to it had never carried information. This is the paper's first result, and it was invisible until someone counted. As a sensitivity check rather than a correction, since neither available variable is exactly the production distribution of the heuristic: reweighting the v2 rows to the production distribution of the stored type, the variable the sample was stratified on, raises the type accuracy of every zero-shot system by 4--10 points and B's by 4 (to 0.828, still the highest) and lowers the stored production type to 0.13; with the keyword heuristic recomputed on every email instead, no system's type accuracy moves by more than 3 points except the stored production type's ($-3.5$).

\textbf{Two hundred labels and a logistic regression have the highest point accuracy on every field, but no significant lead over the best hosted model on any---and never predict the two rarest classes.} Out-of-fold TF-IDF + logistic regression leads in point accuracy on type, urgency, the three switches and project assignment (Tables~\ref{tab:type}, \ref{tab:bool} and~\ref{tab:proj}), in under a millisecond. After Holm correction over the 38 comparisons with B (Table~\ref{tab:tests}) its lead is significant against the stored production type on type, against most local models, and on action-48h against every system but Jev; on every field at least one hosted model is statistically indistinguishable from it. On needs-reply, two blind annotators agree with the reference only at $\kappa=0.15$ and $0.05$ (Section~\ref{sec:limits}). The claim is about accuracy: on type its macro-F1 (0.47) is below Gemma~4's (0.54) because it never predicts \emph{offer} or \emph{spam}, which have fewer threads than folds. Two ablations qualify the win (Table~\ref{tab:abl}). Removing the sender from the baseline's input costs four points on type (0.74 to 0.70 in the ablation harness, whose folds differ slightly from the main run), one to two on the switches and none on urgency: most of the advantage is textual, and the sender matters chiefly for \emph{other}, which is almost entirely the owner's own mail (16/18 rows; no zero-shot system reaches F1 0.3 on it, B gets 0.83). Replacing TF-IDF with a sentence-embedding model (MiniLM) and the same classifier does not help: the English \code{all-MiniLM-L6-v2} is below the majority class on type (0.55) and the multilingual \code{paraphrase-multilingual-MiniLM-L12-v2} reaches 0.69, still below the n-gram baseline, which on this corpus benefits from word bigrams and sender tokens (adding character 3--5-grams moves type from 0.771 to 0.785 over five seeds). Aggregating the three rare types into one class (\emph{other}+\emph{offer}+\emph{spam}, 26 rows) gives a four-class view in which macro-F1 is stable: B 0.57, Tev1 4B 0.58, Nimble 9B 0.50, against 0.57 majority accuracy.

\textbf{Which differences are real.} McNemar's test on paired per-email correctness separates the comparisons that matter from those that do not; the $\chi^2$ values below are uncorrected, and Table~\ref{tab:tests} gives the exact tests against B with Holm correction. On type, B is significantly better than the stored production type ($\chi^2=92$, $p<0.01$) and, before correction only, than Nimble ($\chi^2=4.1$, $p<0.05$; Holm $p=0.51$), but not than Gemma~4 26B or 31B ($\chi^2=1.2$, $1.8$, n.s.): the trained baseline and the best generative model are tied on type. Gemma~4 26B is significantly better than Qwen2.5-7B ($\chi^2=27$) and Nimble than Tev1 0.8B ($\chi^2=4.4$). On urgency, B and Nimble are tied ($\chi^2=0.2$) and both beat the Gemmas before correction ($p<0.05$; for B not after Holm, $p=0.29$ and $0.49$); B's lead over the stored production type on urgency is not significant ($\chi^2=2.6$). Jev and Nimble v4-max are tied on type (exact McNemar $p=0.33$) and Nimble is better on urgency ($p<10^{-6}$); on the switches, asked in the same \code{choice} format, they are tied on needs-reply ($p=0.58$), Jev is better on action-48h and Nimble on is-noise (both $p<0.01$). Asked as \code{noul} instead, Jev's is-noise falls to 0.675 while Nimble's does not move (0.866, $p=0.82$ against its own \code{choice} answers): the gap is the model's, and Jev is the more format-sensitive of the two. Bootstrap intervals are $\pm 5$--$6$ points throughout.

\textbf{Among zero-shot systems the ranking depends on the field.} On type the generative Gemma~4 26B-A4B is best, with the best F1 on the four large classes, and the hosted decision models Jev and Nimble v4-max come next; on urgency and is-noise Nimble is best, and on needs-reply Nimble and Gemma 31B tie in accuracy while Jeva-4B has the highest macro-F1, well inside the intervals (Tables~\ref{tab:type} and~\ref{tab:bool}). Nimble costs about \$0.02 per 1{,}000 emails, Gemma 26B \$0.03. Qwen2.5-7B---the same open weights a user could run locally---is at or below the majority class on both type and urgency. Giving the generative models examples does not close the gap: with three labelled examples per class in the prompt (drawn from each fold's training part), Qwen gains one accuracy point and nine macro-F1 points on type, neither significant ($p=0.77$, $0.10$), and Gemma~4 26B is unchanged (0.74 to 0.73), at six to eight times the cost per email.

\textbf{Local decision models: size matters more than quantised depth.} On a T4, Tev1 4B is the best local decision model on type, above Tev1 0.8B, Jeva-4B and the locally served Nimble 9B in Q8 (Table~\ref{tab:type}); the hosted Nimble v4-max does better than its local build (0.70 vs 0.62), so either the Q8 quantisation or the serving stack costs roughly eight points. All local decision models fail on urgency (0.25--0.45): they assign \emph{medium} or \emph{high} to most messages (Tev1 0.8B: 151 and 66 of 231), where \emph{low} is 70\% of the labels.

\textbf{action-48h is the hard switch.} With 20 positives in 231, no zero-shot model beats the majority class in accuracy (0.91); all beat it in macro-F1 (0.54--0.59 vs 0.48), B reaches 0.73. This is the switch with the highest cost of error---a task created or not---and the one where a classifier needs the most examples.

\textbf{The relationship belongs in the contact graph.} Where a contact has at least one other labelled email (146 of 231, 63\%), inheriting the relationship from the same address is right 98.6\% of the time; the remaining 37\% of emails come from senders seen once, for whom a majority fallback is used (Table~\ref{tab:rel}). In production the covered fraction grows with every labelled contact, and a new sender is exactly the case where the agent should ask rather than guess. Matching on domain covers more but errs more: \code{gmail.com} mixes the owner, clients and partners. As a control we asked three hosted models to infer the relationship (eight classes) from the text of one message, with the same cleaned input and question style as for the other fields: Gemma~4 26B reaches 0.602 (macro-F1 0.28), Nimble v4-max 0.533 and Jev 0.494, against a 0.498 majority class. A generative model recovers part of the relationship from the text; inheritance from the contact gives 0.857 on all emails, with no model call. The agent's production logic now inherits it from the contact (Section~\ref{sec:changes}).

\textbf{Project assignment needs descriptions, not names.} Given only the project names, zero-shot models score 0.26--0.47 and over-use \emph{none} (precision 0.13--0.22); the baseline, which learns from the other emails of a thread's neighbours, reaches 0.70. With a one-line description per project written by the owner (what the project is and what its emails are about, with no names of people or companies), the locally served Nimble 9B doubles its accuracy on the v3 rows with a project, above the majority class, while Tev1 4B stays below it (Table~\ref{tab:projdesc}). Two practical findings came with it: Ollama's \code{/v1/systemone} accepts nineteen options without complaint, but silently fails when the option descriptions are long---a first run with full-sentence descriptions returned an error on every call, and truncating each description to 120 characters fixed it; and 13 of the 352 emails belonged to projects created after the descriptions were written, a reminder that the project list is a moving target maintained by the agent's project-manager role. The descriptions themselves are not released. The hosted models gain the same way, Nimble v4-max most; it almost stops answering \emph{none} (recall 0.15 on the 13 true cases), while Jev and Gemma keep over-using it (precision 0.11 and 0.09). Jeva-4B~\cite{jeva}, with \emph{none} as an ordinary option like every other system, scores 0.114 / 0.191 with descriptions, below the majority class, and 0.210 / 0.216 with names only (its native null is reported separately in Table~\ref{tab:projdesc}). With descriptions the best zero-shot system comes within eight points of the trained baseline on the same 352 rows (0.682 vs 0.761), from 47 points with names only.

\begin{figure*}[t]
\centering
\includegraphics[width=\textwidth]{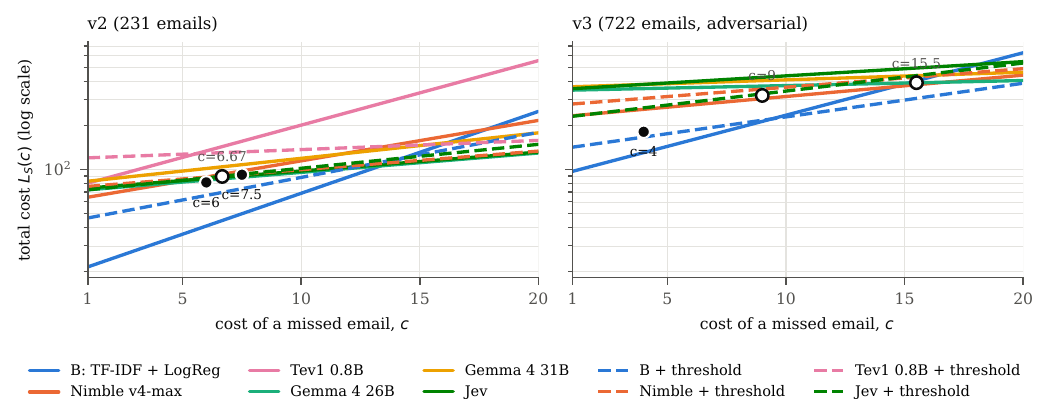}
\caption{Total cost $L_S(c)=c\,m_S+t_S+0.5\,d_S+n_S$ (Eq.~\ref{eq:cost}, $u=1$) of each system's actions (is-noise wins) as the price $c$ of a missed email grows; log scale; each cost is a straight line in $c$. Dashed: the same model with the is-noise threshold chosen inside thread-grouped cross-validation to minimise $L_S(10)$ on each training part (Gemma returns no probability and has no threshold). Filled dot: the cheapest system changes; hollow dots: the same among systems without a threshold. Jev's switches are asked as \code{choice} on v2 and as \code{noul} on v3.}
\label{fig:cost}
\end{figure*}

\begin{table*}[t]
\centering\small
\caption{Composition policies for the three switches, v2 and v3. The truth is fixed (labels composed with \emph{is-noise} wins); only the composition of the predictions changes. \emph{Noise wins}: a predicted is-noise ignores the email (the policy of Table~\ref{tab:actions}); \emph{action wins}: a predicted reply or task overrides is-noise; \emph{consistency}: is-noise together with a reply or a task is not ignored and not acted on, but flagged to the owner. FR = false noise (emails that needed a reply or a task and were ignored), t = useless tasks, d = useless drafts, n = noise emails left in the inbox, f = flagged conflicts, $L_{10}$ = $L_S(10)$ of Eq.~(\ref{eq:cost}) with a flag costing 0.5. Thresholds and Jev formats as in Figure~\ref{fig:cost}; Gemma returns no probability and is not thresholdable as served.}
\label{tab:policy}
\resizebox{\textwidth}{!}{%
\begin{tabular}{l rrrrr rrrr rrrr}
\toprule
 & \multicolumn{5}{c}{Noise wins} & \multicolumn{4}{c}{Action wins} & \multicolumn{4}{c}{Consistency} \\
\cmidrule(lr){2-6}\cmidrule(lr){7-10}\cmidrule(lr){11-14}
System & FR & t & d & n & $L_{10}$ & FR & t & d & $L_{10}$ & FR & t & f & $L_{10}$ \\
\midrule
\multicolumn{14}{l}{\emph{v2 (231 emails, 68 needing a reply or a task)}} \\
B: TF-IDF + LogReg & 12 & 2 & 7 & 4 & 129.5 & 9 & 4 & 7 & 103.5 & 9 & 2 & 5 & 104 \\
B + threshold & 7 & 3 & 7 & 33 & 109.5 & 4 & 4 & 7 & 81.5 & 4 & 3 & 4 & 82.5 \\
Nimble v4-max & 8 & 28 & 19 & 19 & 136.5 & 8 & 30 & 21 & 143.5 & 8 & 28 & 4 & 142.5 \\
Nimble + threshold & 3 & 30 & 19 & 34 & 103.5 & 3 & 30 & 21 & 106.5 & 3 & 30 & 2 & 106.5 \\
Jev & 3 & 17 & 23 & 43 & 101.5 & 3 & 17 & 27 & 105.5 & 3 & 17 & 4 & 105.5 \\
Jev + threshold & 4 & 17 & 23 & 40 & 108.5 & 4 & 17 & 27 & 112.5 & 4 & 17 & 4 & 112.5 \\
Gemma 4 26B-A4B & 3 & 19 & 21 & 40 & 99.5 & 3 & 20 & 22 & 103 & 3 & 19 & 2 & 102.5 \\
Gemma 4 31B & 5 & 21 & 22 & 46 & 128 & 5 & 22 & 23 & 131.5 & 5 & 21 & 2 & 131 \\
Tev1 0.8B & 25 & 11 & 9 & 40 & 305.5 & 22 & 23 & 12 & 298 & 22 & 11 & 14 & 291.5 \\
Tev1 0.8B + threshold & 2 & 22 & 12 & 90 & 138 & 2 & 23 & 12 & 140 & 2 & 22 & 1 & 139.5 \\
\multicolumn{14}{l}{\emph{v3 (722 emails, 120 needing a reply or a task)}} \\
B: TF-IDF + LogReg & 28 & 17 & 26 & 39 & 349 & 23 & 20 & 28 & 305 & 23 & 17 & 9 & 305.5 \\
B + threshold & 13 & 19 & 28 & 96 & 259 & 13 & 20 & 28 & 261 & 13 & 19 & 1 & 260.5 \\
Nimble v4-max & 11 & 58 & 48 & 140 & 332 & 10 & 81 & 61 & 385.5 & 10 & 58 & 37 & 374.5 \\
Nimble + threshold & 11 & 62 & 49 & 183 & 379.5 & 10 & 81 & 61 & 423.5 & 10 & 62 & 32 & 414.5 \\
Jev & 10 & 29 & 54 & 291 & 447 & 9 & 32 & 57 & 446.5 & 9 & 29 & 6 & 445 \\
Jev + threshold & 16 & 26 & 48 & 165 & 375 & 15 & 32 & 57 & 388.5 & 15 & 26 & 15 & 385.5 \\
Gemma 4 26B-A4B & 3 & 77 & 88 & 225 & 376 & 3 & 86 & 99 & 408.5 & 3 & 77 & 19 & 403.5 \\
Gemma 4 31B & 5 & 84 & 76 & 241 & 413 & 5 & 96 & 87 & 452.5 & 5 & 84 & 23 & 446.5 \\
\bottomrule
\end{tabular}}
\end{table*}

\textbf{Accuracy is not the product metric.} The labels compose into one of four actions per email: draft a reply, create a task, ignore, or nothing, with the rule that \emph{is-noise} wins. Scored that way (Table~\ref{tab:actions}), the error that costs most---\emph{false noise}, an email that needed a reply or a task and was ignored---inverts the ranking. On v2 the trained baseline ignores 18\% of the 68 emails that needed action, against 4\% for Gemma~4 26B; on v3, B 23\% against Nimble 9\%. The baseline is precise and conservative, the LLMs recall more and over-act. The same dissociation between reading and acting is reported at scale by Li et al.~\cite{li26}: on 19{,}663 frozen agent states, 73--85\% of the decision errors of the strongest System One models fall on states whose facts the model reads correctly, and the share grows with accuracy; their remedy---ask the facts as narrow questions and compose the action in code---is the design our three switches already follow, and their confidence cascade, which re-asks only below a threshold~\cite{li26judge}, is the natural next step for the noise decision here, where most tasks the zero-shot models create are useless (Table~\ref{tab:actions}). Inheriting the relationship from the contact and applying the monitoring rule raises task recall for every system (B 0.25 to 0.40, Gemma 0.25 to 0.65) at no cost in false noise; the oracle relationship adds nothing more, so the relationship is not the bottleneck. The composition rule matters too: letting the reply and task switches win over is-noise (\emph{action wins}) recovers 3 of the baseline's 12 false-noise emails on v2 and 5 of 28 on v3 without changing the model, and flagging the conflict to the owner instead (\emph{consistency}) recovers the same emails without creating any extra task, at the price of 5 and 9 flagged emails (Table~\ref{tab:policy}). The idea of composing narrow decisions in code comes from Li et al.~\cite{li26}; applying it to the noise switch was suggested by Yubo Li (personal communication). Which system is right depends on the price of a missed email versus a useless task. Making that price explicit settles it. For a system $S$ let $m_S$ be the number of emails that needed a reply or a task and were ignored, $t_S$ the tasks created for emails that needed none $d_S$ the drafts prepared for emails that needed none and $n_S$ the noise emails left in the inbox; with unit costs $c$, $1$, $0.5$ and $u$ the total cost is
\begin{equation}
L_S(c) \;=\; c\,m_S \;+\; t_S \;+\; 0.5\,d_S \;+\; u\,n_S ,
\label{eq:cost}
\end{equation}
and the system to deploy is $\arg\min_S L_S(c)$ for the owner's $c$. We set $u=1$: with $u=0$ a threshold chosen for cost degenerates---the baseline then ignores 109 of the 231 emails, for a cost $L_S(10)$ of 7.5---so the noise left in the inbox must have a unit price, which is also what the 10:1 rule of the earlier versions of this paper did implicitly. Figure~\ref{fig:cost} plots $L_S(c)$ for $c$ from 1 to 20. Among the systems without a threshold the baseline is cheapest while $c<6.7$ on v2 ($c<9$ on v3); above that Gemma~4 26B is cheaper on v2, and on v3 Nimble v4-max and then, from $c=15.5$, Gemma~4 26B; at $c=10$ on v2 the baseline costs 129.5, against 101.5 for Jev and 99.5 for Gemma~4 26B. For a personal agent, where a missed client email is worth far more than a task to dismiss, that favours the zero-shot models out of the box. But the baseline's false-noise rate is a threshold, not a ceiling: choosing the \emph{is-noise} threshold inside the cross-validation (on each training part, the threshold that minimises $L(10)$) cuts it from 12/68 to 7/68 on v2 and from 28/120 to 13/120 on v3 (Table~\ref{tab:policy}). With that threshold the baseline's false noise is no longer distinguishable from Nimble's (v2: 7/68 vs 8/68, exact McNemar $p=1.0$; v3: 13/120 vs 11/120, $p=0.77$), and its cost $L_S(10)$ falls from 129.5 to 109.5 on v2 and from 349 to 259 on v3, while it keeps its accuracy lead. The same threshold given to Nimble v4-max cuts its false noise to 3/68 on v2 (cost 136.5 to 103.5) but not on v3 (11/120, cost 332 to 379.5). On v2 no cost difference among the leading systems is significant (thread bootstrap at $c=10$, thresholded baseline minus Gemma~4 26B: $+10$, 95\% CI $[-57, 81]$; minus Jev: $+8$, $[-58, 82]$); the cheapest system is the baseline up to $c=6$, the thresholded baseline up to $c=7.5$ and Gemma~4 26B above. On v3 the thresholded baseline is the cheapest system for every $c>4$, and at $c=10$ it is cheaper than every other system (95\% CIs exclude zero for all but Nimble v4-max, $[-151, 9]$). Choosing the threshold again for each $c$ instead of at $c=10$ moves the v2 crossover to Gemma from 7.5 to 7 and the v3 one from 4 to 5. Calibration itself is not what does the work---a cost-chosen threshold depends only on the ranking of probabilities---but it is what makes the probabilities safe to display or combine, the precondition for using confidence to control decisions~\cite{li26survey}.

\textbf{Calibration.} Nimble v4-max is the best-calibrated system on type, urgency and action-48h and well calibrated on every field (ECE 0.04--0.11 on v2, 0.04--0.09 on v3); on needs-reply and is-noise Tev1 0.8B (0.07, 0.04) and Jev asked as \code{noul} (0.06, 0.06) are slightly better calibrated on v2. Nimble is deterministic (491/491 identical answers on a repeated run). The trained baseline is strongly \emph{under}-confident: on type its mean confidence is 0.32 for an accuracy of 0.79 (ECE 0.47), an artefact of balanced class weights on few rows; recalibrating it inside the cross-validation (Platt or isotonic, fitted on each outer fold's training part) cuts its ECE by a factor of 4--17 (type: 0.47 to 0.11 on v2, 0.37 to 0.05 on v3; is-noise: 0.29 to 0.03), to the same range as Nimble v4-max, at a cost of five accuracy points on type when the corpus is small (231 rows) and none when it is larger (722). The local decision models are over-confident on errors (Tev1 4B 0.81 right vs 0.70 wrong on type; Nimble 9B Q8 0.88 vs 0.76).

\textbf{The adversarial corpus is a stress test, not a second sample.} On v3 (722 emails) the baseline's lead over Gemma~4 26B on type becomes significant (+0.136, $p<0.001$), as does its lead over Nimble on every switch; the contact-graph relationship reaches 0.927 on the 73\% of covered emails and 0.898 overall (coverage is inflated on v3, where G3 was drawn by sender strata and many senders recur: the v2 figures, 63\% covered and 0.857 overall, are the ones to carry to production); the stored production type falls to 0.17 on type; the local decision models on a T4 fall below the majority class on every field (Tev1 4B 0.52 on type, Nimble 9B Q8 0.50; 0.40 and 0.34 on urgency), and so does the hosted Nimble v4-max on type (0.56 against 0.58), while Jev (0.60) and Gemma~4 26B (0.65) stay above it. But v3 was sampled on the LLMs' presumed errors, and the baseline was the first proposal shown to the reviewer, so these gaps are upper bounds, not estimates. The pre-registered hypotheses give the honest summary: seven of eleven confirmed, four refuted---among the refuted, that rare commercial classes could be found by sampling where the LLMs see them (all 118 such emails yielded 3 \emph{offer} and 7 \emph{spam}: the models over-predict commercial mail, the inbox simply has little of it after Gmail's own filtering), and that the baseline would learn \emph{offer} with eight examples (its F1 stays 0).

Jev (TypeSafe, served as jev-1.14.0), the reference model of the category, was added in this version with the same Italian questions (Tables~\ref{tab:type} and~\ref{tab:bool}). It agrees with Nimble v4-max on 86\% of type labels, is the fastest hosted system (0.22\,s median per decision, 0.29\,s p95), costs \$0.11 per 1{,}000 emails for all five questions, and is not deterministic: on 100 repeated emails its label is unchanged in 97--100 cases per field but its probabilities in only 15--40; its ECE on v2 is 0.12--0.14 in the \code{choice} format and 0.06--0.08 on the switches asked as \code{noul}. Jev was not run in v1 for lack of an API key; there the conclusions about typed decision models rested on four of them (Tev1 0.8B and 4B, Nimble 9B local, Nimble v4-max hosted), and Li et al.~\cite{li26} report Jev and Nimble-class models failing in the same way on the same states, so the architectural finding---read with narrow questions, compose in code---does not depend on the proprietary reference; Nimble's public demo was offline and the paid API was used; GPU runs were made on Kaggle's free T4 with the same script and data. Jeva-4B~\cite{jeva} (Qwen3.5-4B backbone), added in v2.1 and run on a Kaggle T4 with the same questions, is below the majority class on type (0.498 vs 0.571), on urgency (0.264) and on project assignment (Table~\ref{tab:projdesc}); these low scores may reflect genuine limits of zero-shot use on a multilingual and highly personal inbox (Section~\ref{sec:limits}). The part of its design that this inbox exercises holds: on the three switches asked as \code{choice} it is level with the hosted systems (Table~\ref{tab:bool}), and reusing option encodings gives identical decisions on 352/352 rows at 1.03\,s per call against 3.6--5.3\,s without the cache, a third to a fifth of the latency. It is not in the sensitivity, action, cost or annotator analyses.

\section{What Changed in Production}\label{sec:changes}
The measurements changed the system. Between the second and third labelling round the production pipeline was rewritten to follow the label design of Section~\ref{sec:setup}: (i) type and urgency are no longer produced by the keyword heuristic but by the configured classifier, with the heuristic kept only as a logged fallback; (ii) the sender's relationship is a property of the contact, inherited by every email from that address, never asked of a model; (iii) monitoring notifications are routed by a rule---a problem creates a task, an all-clear is kept as context; (iv) the three switches are the only triggers of the agent's actions (draft, task, ignore), replacing earlier conditions that mixed type and urgency; (v) private mail marked \emph{ignore} leaves the flow entirely; (vi) project names are proposals of the project-manager agent and may be renamed or merged. Each new decision passes through the same logging path as the ones measured here, so the next round can be measured the same way. In the first six days after the change (22--27 October 2026) 129 emails went through the new pipeline: 102 were classified by the generative LLM configured as classifier (the decision models of this paper are not yet wired into production) and 27 by rules (monitoring, ignore), none by the keyword fallback; the switches asked for 7 drafts and 23 tasks (22 of them from the monitoring rule), of which 3 drafts and 14 tasks were created, and 63 emails were ignored as noise.

\section{Error Analysis}\label{sec:errors}

From the divergences between the six systems we counted, per cause, the emails on which at least three systems err in the direction of that cause (\code{ERRORI.md}).
\begin{enumerate}
\item \textbf{Relational classes.} In the first round all 5 \emph{client} emails were misread as \emph{request} or \emph{update} by at least three systems; after the label change, \emph{other} (mostly the owner's own mail) plays the same role: no zero-shot system reaches F1 0.3 on it. Who the sender is to the owner is not in the text of one message; it lives in the contact graph (Table~\ref{tab:rel}).
\item \textbf{Automated senders are not the problem.} We expected deploy, CI and console notifications to be misfiled; on the reviewed labels no email with an automated sender is misclassified by any system. The rule-based robot filter in front of the models already removes the hard cases, and the models handle the rest.
\item \textbf{Thin text.} Bodies under 20 characters are now excluded as \emph{ignore}; in the first round every system failed on them on both fields.
\item \textbf{Debatable labels.} On 9 type labels and 5 urgency labels at least four of the six systems agree with each other against the reviewer. These are candidates for independent annotation rather than evidence of model error; at 8--12\% of the corpus they bound the achievable accuracy. Three of the nine contested type labels fall in the blind annotators' 60 emails: the first annotator agrees with the reference on two and with the systems on none, and gives a third label on the last; the second agrees with the systems on one and gives a third label on the other two; none of the five contested urgencies is in that sample.
\item \textbf{Rare commercial classes are rare.} Sampling every email in which two LLMs saw \emph{offer} or \emph{spam} (118 of a 1{,}322-email pool) produced 3 and 7 true cases: Nimble's precision on \emph{spam} is 0.10. The models over-predict commercial mail; the inbox, filtered upstream by Gmail, has little of it.
\item \textbf{Taxonomy drift.} The stored \emph{type} mixes message kind (request, update, offer) with relationship (client, spam), which is the root of cause 1.
\end{enumerate}

\section{Limitations}\label{sec:limits}

\begin{table}[t]
\centering\small
\caption{Two blind annotators (A1, A2) on the same 60 v2 emails: Cohen's $\kappa$ of each against the reference labels and between them, and Fleiss' $\kappa$ of the three, with 95\% bootstrap intervals (2{,}000 resamples). Last column: yes (switches) or majority class (type, urgency) for reference / A1 / A2.}
\label{tab:kappa}
\resizebox{\columnwidth}{!}{%
\begin{tabular}{lccccc}
\toprule
Field & A1 vs ref. & A2 vs ref. & A1 vs A2 & Fleiss (3) & ref./A1/A2 \\
\midrule
type & 0.81 [0.68, 0.93] & 0.86 [0.74, 0.95] & 0.81 [0.68, 0.92] & 0.83 [0.73, 0.91] & update 28/29/28 \\
urgency & 0.63 [0.44, 0.81] & 0.44 [0.25, 0.62] & 0.66 [0.49, 0.83] & 0.57 [0.41, 0.71] & low 43/41/32 \\
needs-reply & 0.15 [$-$0.06, 0.39] & 0.05 [$-$0.14, 0.28] & 0.35 [$-$0.07, 0.73] & 0.11 [$-$0.08, 0.32] & yes 18/5/5 \\
action-48h & 0.50 [$-$0.03, 0.85] & 0.35 [$-$0.07, 0.68] & 0.70 [0.27, 1.00] & 0.51 [0.06, 0.78] & yes 7/4/7 \\
is-noise & 0.67 [0.46, 0.83] & 0.37 [0.12, 0.60] & 0.57 [0.34, 0.77] & 0.53 [0.36, 0.69] & yes 31/31/32 \\
\bottomrule
\end{tabular}}
\end{table}

Single user, single inbox, two languages. What would not transfer to another user is clear from the components: the baseline's weights (they encode this owner's senders and vocabulary and must be retrained from that user's labels), the contact graph (empty for a new user until contacts recur, which is why coverage, not accuracy, is the quantity to watch), the class priors (70--82\% \emph{low} here) and any confidence threshold. What should transfer is the protocol itself, the zero-shot behaviour of the decision models, and the finding that the sender's relationship is a graph property; the second inbox needed to test this is the obvious next study; the owner reviewed labels proposed by model annotators rather than labelling from scratch (110/115 first-round type proposals survived review), and the final label definitions were applied in a joint revision with an assistant, which may reflect anchoring as much as agreement (two independent blind annotators on 60 emails, below). 231 labels give 95\% confidence intervals of roughly $\pm$6 points on accuracy, so differences below eight points between systems are not significant; the gap between the production heuristic and every other system, and between the trained baseline and the zero-shot models on is-noise, are. The sample was stratified on the production heuristic's prediction, so its accuracy is not a production estimate. The data cannot be released; the protocol, the harness and aggregate statistics can. Round 3 was sampled on the LLMs' presumed errors and shown with the baseline's proposal first; v3 comparisons inherit both biases. Two blind annotators, who saw only sender, subject, date and body, labelled the same 60 v2 emails independently (Table~\ref{tab:kappa}). Agreement with the reference is almost perfect on type for both ($\kappa=0.81$, $0.86$; Fleiss' $\kappa$ of the three 0.83), moderate to substantial on urgency and is-noise (0.44--0.67), and weak on needs-reply (0.15, 0.05), where each annotator marked 5 emails against the reference's 18; the two annotators agree with each other more than with the reference on needs-reply (0.35). Needs-reply therefore encodes the owner's own threshold for what deserves an answer more than a property the text settles, and action-48h, with 7 positives, cannot be judged on this sample (intervals from below zero to 0.68--1.00). Scored against either blind annotator instead of the reference on the same 60 emails, the ranking on type is unchanged (the two hosted decision models first, the baseline fifth or sixth on this subset), but the baseline loses its first place on urgency and on the three switches (second to sixth): part of its advantage is that it learns the owner's conventions, which no zero-shot model can know. The two gaps left open in v1---a reweighting of the zero-shot accuracies by the production distribution of the stratification variable, and a project-assignment run with a one-line description per project written by the owner, since names alone are not a fair zero-shot input (Table~\ref{tab:proj})---are now reported in the Results. Few-shot prompting and recalibration of the baseline, raised by early reviewers, have been run and are reported above. The generative naming step for new projects was run through a hosted endpoint serving open weights (Qwen2.5-7B-Instruct) that are deployable locally; local latency for that step was measured only on a small subsample. Local GPU results come from Kaggle's free T4, not from the owner's hardware; the locally served Nimble is the Q8 build. Jeva-4B was trained mostly on English decision data (the authors, personal communication, November 2026), while questions and options were in Italian for every system (Appendix~B), so its input is out of distribution; ablations with English questions are possible but were not run. The hosted generative model was not deterministic across runs at temperature 0 (0.774--0.800 on type), so hosted figures should be read as single-run estimates.

\section{Conclusion}

Two days of measurement on 231 emails and eleven systems showed that the field a personal agent displayed most confidently had never carried information; that two hundred human labels and a logistic regression reach the highest point accuracy on every one of the agent's decisions, at negligible cost, but without a significant lead over the best hosted model on any of them; that among zero-shot options a hosted typed decision model is the best choice for the agent's switches and a mid-size generative model for message type, with Tev1 4B, a local decision model on a consumer GPU, close behind; that the sender's relationship, which a generative model recovers only in part from one message (0.60), is recovered almost perfectly from the contact graph where the sender is known; and that the most accurate system ignores an email that needed action more often than any hosted model in Table~\ref{tab:actions}, so the choice between a classifier and a model is a choice about the price of errors, not about accuracy---and once that price is written down, a threshold chosen for it cuts the classifier's missed emails by about half (12 to 7 of 68 on v2, 28 to 13 of 120 on v3), to the level of the hosted decision model; on the unbiased corpus no cost difference among the leading systems is significant, and on the adversarial one the thresholded classifier is the cheapest system above $c=4$. The protocol---stratified sampling, frozen snapshots, thread-grouped splits, secret redaction, counting before training---cost less than the labelling and is reusable for any personal agent. The production system was changed accordingly (Section~\ref{sec:changes}): type and urgency are no longer rule-based, the relationship is inherited from the contact, and the three switches drive the agent's actions directly.

\paragraph{Reproducibility.} Code and protocol: \url{https://github.com/federicolucca/foyer-measure/tree/v2.3}. The public harness contains the questions and definitions used here (\code{systemone\_email.py}), the benchmark and the hosted backends (\code{bench\_email.py}, \code{decision\_bench.py}, \code{online\_backends.py}), the Jeva-4B backend (\code{jeva\_backend.py}), the metrics (\code{metrics.py}), the cost-chosen threshold (\code{cost\_threshold.py}), project assignment (\code{project\_bench.py}) and inter-annotator agreement (\code{kappa\_multi.py}), with a README listing every command and environment variable name. The scripts that read the production database, the run log and the per-email result files stay private with the data. GPU runs: a single-cell Kaggle notebook (T4) with the same questions. Models: \code{tev1:0.8b}, \code{tev1}, \code{nimble} on Ollama 0.40; scikit-learn 1.9.

\paragraph{Acknowledgements.} We thank Haithem Afli (Munster Technological University) for his support and for endorsing the arXiv submission, and Alessandro Resoconto and Davide D'Agostini for the blind annotation of 60 emails, and Qi Li for confirming the Jeva setup.

\paragraph{Changes in this version (v2.2).} Corrections after public reviews of v2.1. (i) The is-noise threshold and cost results of v2.1 had been computed on a copy of the corpus with bodies truncated to 700 characters. Recomputed on the benchmark corpus (bodies up to 1{,}500 characters; the baseline's is-noise predictions match the saved run on 231/231 and 718/722 rows), the thresholded baseline's false noise is 7/68, not 4/68, and it is no longer the cheapest system at every price; the cost now prices the noise left in the inbox ($u=1$ in Eq.~\ref{eq:cost}), crossovers are exact and the cost differences carry thread-bootstrap intervals (Figure~\ref{fig:cost}, Table~\ref{tab:policy}). (ii) The baseline's features are word 1--2-grams, as run, not word and character n-grams; it is reported over five seeds of the split. (iii) Thread-cluster intervals and Holm-corrected exact McNemar tests for every system (Tables~\ref{tab:type} and~\ref{tab:tests}); the baseline's accuracy lead is restated as point accuracy without a significant lead over the best hosted model. (iv) The ``production rules'' column is the stored production type, with its provenance. (v) One definition of \emph{client}. (vi) Sentences whose source runs were not saved (local rephrasing and language experiments, a local relationship control) were removed; the relationship control was re-run on hosted models. Code and revision scripts: tag v2.3.

\appendix
\begin{table*}[t]
\centering\small
\caption{All systems on v2 (231 emails): accuracy with 95\% thread-cluster bootstrap intervals (2,000 resamples of the 207 threads). $^{*}$: exact McNemar test against B significant at 0.05 after Holm correction over all 38 comparisons in the table; $(n_{B}{:}n_{S})$ = emails only B / only the system got right. B is the seed-0 run (over five seeds of the split, type $0.771\pm0.021$). Empty cells: field not asked.}
\label{tab:tests}
\resizebox{\textwidth}{!}{%
\begin{tabular}{lccccc}
\toprule
System & Type & Urgency & Needs-reply & Action-48h & Is-noise \\
\midrule
B: TF-IDF + LogReg & 0.788 [.726, .846] & 0.736 [.667, .802] & 0.831 [.769, .886] & 0.931 [.890, .966] & 0.900 [.854, .942] \\
Stored production type & 0.277 [.219, .339] (133:15)$^{*}$ & 0.684 [.613, .752] (29:17) &  &  &  \\
Tev1 0.8B & 0.623 [.556, .687] (64:26)$^{*}$ & 0.251 [.191, .317] (144:32)$^{*}$ & 0.706 [.634, .770] (40:11)$^{*}$ & 0.840 [.788, .889] (27:6)$^{*}$ & 0.628 [.564, .692] (69:6)$^{*}$ \\
Tev1 4B & 0.688 [.622, .754] (56:33) & 0.450 [.383, .522] (101:35)$^{*}$ &  &  &  \\
Nimble 9B Q8 & 0.619 [.549, .685] (69:30)$^{*}$ & 0.411 [.346, .478] (109:34)$^{*}$ &  &  &  \\
Jeva-4B & 0.498 [.428, .567] (91:24)$^{*}$ & 0.264 [.204, .329] (130:21)$^{*}$ & 0.792 [.732, .848] (36:27) & 0.805 [.742, .867] (38:9)$^{*}$ & 0.732 [.673, .789] (52:13)$^{*}$ \\
Nimble v4-max & 0.701 [.635, .763] (54:34) & 0.714 [.646, .778] (41:36) & 0.805 [.748, .859] (30:24) & 0.805 [.743, .867] (38:9)$^{*}$ & 0.857 [.803, .903] (25:15) \\
Jev & 0.727 [.664, .789] (47:33) & 0.550 [.479, .618] (75:32)$^{*}$ & 0.792 [.735, .846] (35:26) & 0.857 [.802, .909] (25:8) & 0.783 [.722, .838] (48:21)$^{*}$ \\
Qwen2.5-7B & 0.571 [.502, .639] (78:28)$^{*}$ & 0.437 [.365, .510] (105:36)$^{*}$ &  &  &  \\
Gemma 4 26B-A4B & 0.740 [.683, .801] (46:35) & 0.628 [.564, .692] (66:41) & 0.796 [.743, .846] (32:24) & 0.849 [.791, .898] (28:9)$^{*}$ & 0.809 [.753, .862] (42:21) \\
Gemma 4 31B & 0.732 [.665, .795] (47:34) & 0.645 [.577, .716] (57:36) & 0.805 [.751, .858] (31:25) & 0.840 [.782, .891] (30:9)$^{*}$ & 0.771 [.709, .829] (49:19)$^{*}$ \\
\bottomrule
\end{tabular}}
\end{table*}

\section{Fixed-split results}
The fixed thread-grouped split (train 159 / validation 39 / test 33) is recorded in the labelled JSONL and in the repository's result files; it is not used for any figure in this paper because the test split holds no \emph{high} and a single \emph{offer}. It is retained so that future labelling rounds can report a held-out number once the small classes have enough examples.

\section{Questions, definitions and prompts (verbatim)}
All systems used the same Italian question and definitions, from \code{bench\_tev1\_pilota.py}; English glosses in brackets.

\paragraph{B.1 Type.} Question: \emph{``Che tipo di messaggio è questa email per il suo destinatario?''} [What kind of message is this email for its recipient?]
\begin{itemize}
\item \textbf{richiesta}: chi scrive chiede al destinatario qualcosa, o serve una sua azione o risposta [the sender asks something of the recipient, or an action or reply is needed]
\item \textbf{aggiornamento}: informazione, notifica, conferma o esito senza richiesta (codici di sicurezza, avvisi di accesso, prenotazioni, report di stato, thread in copia) [information, notification, confirmation or outcome with no request]
\item \textbf{offerta}: proposta commerciale mirata al destinatario (preventivo, offerta personale) [a commercial proposal aimed at the recipient]
\item \textbf{cliente}: corrispondenza con un cliente anche potenziale su un incarico professionale [correspondence with a client, including a prospective one, about a professional engagement]
\item \textbf{newsletter}: invio di massa periodico o promozionale da un brand o servizio [periodic or promotional mass mailing]
\item \textbf{spam}: commerciale non richiesto da mittente sconosciuto, o phishing [unsolicited commercial mail from an unknown sender, or phishing]
\item \textbf{altro}: tutto il resto, compresi i messaggi inviati dal destinatario stesso [everything else, including messages sent by the recipient]
\end{itemize}
The \emph{cliente} option was dropped after the first round, when relationship was separated from type (Section~\ref{sec:setup}).

\paragraph{B.2 Urgency.} Question: \emph{``Quanto è urgente agire su questa email?''} [How urgent is it to act on this email?] \textbf{alta}: serve un'azione entro circa 48 ore o c'è un appuntamento in giornata [action within about 48 hours, or an appointment today]; \textbf{media}: serve un'azione o una lettura attenta senza scadenza vicina [action or careful reading, no near deadline]; \textbf{bassa}: nessuna azione richiesta [no action required].

\paragraph{B.3 Decision-model request.} One \code{choice} question per call: \code{\{"model": m, "state": \{mittente, oggetto, corpo\}, "questions": \{campo: \{"type": "choice", "instructions": DOMANDA[campo], "criteria": CRITERI[campo]\}\}\}} against \code{/v1/systemone} (Ollama locally, Bespoke's API for Nimble). Body cleaning: style/script blocks, tags and URLs removed, whitespace collapsed, truncated to 1{,}500 characters.

\paragraph{B.4 Generative prompt (OpenRouter).} System message = question + \emph{``Scegli esattamente una etichetta. Definizioni:''} + the definitions above, one per line + \emph{``Rispondi SOLO con JSON: \{"tipo": "<etichetta>"\}''}; user message = the cleaned email as JSON; \code{temperature} 0, \code{max\_tokens} 60, \code{response\_format} \code{json\_object}. Answers outside the vocabulary were counted as errors (1 of 230 calls for Qwen, 0 for Gemma).

\paragraph{B.5 Production heuristic (\code{sync\_gmail.\_classify\_email}).} Lower-cased subject + snippet. Type: \emph{offerta} if any of \{offerta, preventivo, price, quote, proposta\}; else \emph{richiesta} if any of \{richiesta, request, help, supporto, assistenza\}; else \emph{aggiornamento} if any of \{aggiornamento, update, status, ticket, issue\}; else \emph{altro}. Urgency: \emph{alta} on \{urgente, urgent, asap, subito, critico, critical, scadenza, deadline, entro oggi, entro domani, immediate, priorità alta, high priority, bloccante, blocking\}; \emph{media} on \{importante, important, soon, presto, entro la settimana, this week, priorità media, follow up, follow-up, in attesa\}; else \emph{bassa}. The function's own docstring reads ``classificazione euristica semplice---da sostituire con Ollama quando disponibile'' [simple heuristic classification---to be replaced by Ollama when available].

\section{Labelling procedure and guidelines}
\paragraph{Procedure.} A review spreadsheet with one row per email and the columns: id, split, sender, subject, body (700 characters, whitespace normalised), date, first proposal (type, urgency, confidence, note), second proposal (type, urgency), disagreement, confirm, corrected type, corrected urgency, area, project, needs-reply, action-48h, is-noise, skip, reviewer notes. Rows were ordered with disagreements and low-confidence proposals first, then by id. The production prediction was deliberately not shown. A row counts as reviewed when \emph{confirm} is set or a corrected value is given. Four rows were marked \emph{skip} because the reviewer found them ambiguous, and were excluded. The reviewer went through the rows in the spreadsheet order (disagreements first) and reports filling the optional columns (area, project, needs-reply, action-48h, is-noise) on every row; the project column did not survive the import into the label table in this round and is being re-imported. The reviewer applied \emph{client} more narrowly than the first-round written guideline, as ``someone who gives me work'', i.e.\ an actual paying client rather than a prospective one; this is the definition used throughout the paper (Section~\ref{sec:setup}).

\paragraph{Guidelines given to the model annotator (translated).} \emph{request}: the sender asks me for something, or my action or reply is needed (including an automatic notice that says ``action required''). \emph{update}: information, notification, confirmation or outcome with no request: bookings, security codes, login alerts, updated calendar invitations, status reports, legal or pricing notices from a service, threads seen in copy. \emph{offer}: a commercial proposal aimed at me. \emph{client}: correspondence with a client (including prospective) about an ongoing professional engagement. \emph{newsletter}: periodic or promotional mass mailing from a brand or service I subscribed to. \emph{spam}: unsolicited commercial mail from a sender unrelated to me. \emph{other}: the rest, including messages sent by me. Urgency: \emph{high} action within about 48 hours or an appointment today; \emph{medium} action or careful reading without a near deadline (security notices, time confirmations, failed builds); \emph{low} no action. The annotator saw sender, subject and the first 650 characters of the body, without the production prediction; for empty or markup-only bodies it decided from subject and sender.


\begin{thebibliography}{9}\small\setlength{\itemsep}{0pt}\setlength{\parsep}{0pt}
\bibitem{jev} D.~Almeida. Introducing System One Models \& Jev. TypeSafe AI blog, 15 Sept.\ 2026.
\bibitem{nimble} Bespoke Labs, S.~Pimpalgaonkar, P.~Dalal, M.~Sathiamoorthy. Nimble: data, model, recipe for an open Jev. \url{https://github.com/bespokelabsai/nimble}, 2026.
\bibitem{ollama} Ollama. Ollama now supports Jev-style decision models. Blog, 29 Sept.\ 2026.
\bibitem{jeva} Q.~Li, X.~Wang. Jeva: System One Models from First Principles. arXiv:2610.18773, 2026.
\bibitem{llmasjev} Y.~Li, J.~Wagle. LLM-as-Jev: LLMs are already Jev-style decision models---when and how to fine-tune them. arXiv:2610.02076, 2026.
\bibitem{audit} L.~Tang, Y.~Zheng. Typed Decision Models: An Early Evidence Audit and Evaluation Checklist. arXiv:2609.32160, 2026.
\bibitem{bench} X.~Li, Q.~Chang, J.~Ning, C.~Xu, S.~Zhang, Y.~Zhang, L.~Luo, H.~Lin. Specialized Decision Models vs.\ General-Purpose LLMs: Benchmarking Jev Across Knowledge, Reasoning, and Multilingual Tasks. arXiv:2610.11978, 2026.
\bibitem{traffic} S.~Xu, L.~Mei. A First Glance at Jev for Network Traffic Classification: Accuracy, Processing Time, and Cost. arXiv:2610.00376, 2026.
\bibitem{wild} G.~Ling, M.~Xue, Z.~Ye. Jev in the Wild: A Data-Driven Analysis of the Jev Model's Functionality, Applications and Ecosystem. arXiv:2609.30216, 2026.
\bibitem{li26} Y.~Li, R.~Krishnan, R.~Padman. Reading right, acting wrong: diagnosing decision failures of System One models in agent loops. arXiv:2610.18579, 2026.
\bibitem{li26judge} Y.~Li, Y.~Miao, R.~Krishnan, R.~Padman. JEV-as-a-judge: accept when confident, escalate when unsure. arXiv:2609.26550, 2026.
\bibitem{li26survey} Y.~Li, T.~Zhou, X.~Shen, Y.~Miao, R.~Padman, R.~Krishnan. Confidence as control: a survey of confidence utilization in large language models. Preprint, Carnegie Mellon University, 2026. \url{https://yubol-bobo.github.io/assets/pdf/Conf_Survey.pdf}.
\bibitem{cohen96} W.~W. Cohen. Learning rules that classify e-mail. \emph{AAAI Spring Symposium on Machine Learning in Information Access}, 1996.
\bibitem{klimt04} B.~Klimt, Y.~Yang. The Enron corpus: a new dataset for email classification research. \emph{ECML}, 2004.
\bibitem{bekkerman04} R.~Bekkerman, A.~McCallum, G.~Huang. Automatic categorization of email into folders: benchmark experiments on Enron and SRI corpora. CIIR Technical Report IR-418, University of Massachusetts Amherst, 2004.
\bibitem{dredze08} M.~Dredze, T.~Brooks, J.~Carroll, J.~Magarick, J.~Blitzer, F.~Pereira. Intelligent email: reply and attachment prediction. \emph{Proc.\ IUI}, Gran Canaria, 2008.
\bibitem{grbovic14} M.~Grbovic, G.~Halawi, Z.~Karnin, Y.~Maarek. How many folders do you really need? Classifying email into a handful of categories. \emph{Proc.\ CIKM}, pp.\ 869--878, 2014. doi:10.1145/2661829.2662018.
\bibitem{mcmahan17} B.~McMahan, E.~Moore, D.~Ramage, S.~Hampson, B.~Agüera y Arcas. Communication-efficient learning of deep networks from decentralized data. \emph{AISTATS}, 2017.
\end{thebibliography}
\end{document}